\documentclass[letterpaper, 10 pt, conference]{ieeeconf}  % Comment this line out if you need a4paper

\IEEEoverridecommandlockouts                              % This command is only needed if 
\usepackage{graphics} % for pdf, bitmapped graphics files
\usepackage{epsfig} % for postscript graphics files
\usepackage{mathptmx} % assumes new font selection scheme installed
\usepackage{times} % assumes new font selection scheme installed
\usepackage{amsmath} % assumes amsmath package installed
\usepackage{amssymb}  % assumes amsmath package installed
\usepackage{booktabs}
\usepackage[nolist,nohyperlinks,printonlyused]{acronym}
\usepackage{stix}
\usepackage{url}
\usepackage[hidelinks]{hyperref}

\title{\LARGE \bf
Control of Unknown Quadrotors from a Single Throw
}

\author{Till M. Blaha$^1$, Ewoud J.J. Smeur and Bart D.W. Remes% <-this % stops a space
\thanks{+This work has been submitted to IROS 2024 for possible publication. Copyright may be transferred without notice, after which this version may no longer be accessible.}%
\thanks{*This work was not supported by any organization}% <-this % stops a space
\thanks{$^{1}$All authors are with the Faculty of Aerospace Engineering,
        Delft University of Technology, Kluyverweg 1, 2629 HS Delft, Netherlands\newline
        {Correspondence: \tt\small t.m.blaha@tudelft.nl}}
\thanks{Data: \href{https://doi.org/10.4121/0530be90-cc6c-4029-9774-670657882906}{10.4121/0530be90-cc6c-4029-9774-670657882906}}
\thanks{Code: \href{https://github.com/tudelft/indiflightSupport/tree/iros_imav_2024}{github.com/tudelft/indiflightSupport/tree/iros\_imav\_2024}}
}

\begin{document}

\begin{acronym}
\acro{ML}{Machine Learning}
\acro{INDI}{Incremental Nonlinear Dynamic Inversion}
\acro{NDI}{Nonlinear Dynamic Inversion}
\acro{UAV}{Unmanned Air Vehicle}
\acro{RLS}{Recursive Least Squares}
\acro{LMS}{Least Mean Squares}
\acro{ESC}{Electronic Speed Control}
\acro{RMS}{Root-Mean-Square}
\acro{EKF}{Extended Kalman Filter}
\acro{IMU}{Inertial Measurement Unit}
\acro{MRAC}{Model-Reference Adaptive Control}
\end{acronym}

\maketitle
\thispagestyle{empty}
\pagestyle{empty}

%%%%%%%%%%%%%%%%%%%%%%%%%%%%%%%%%%%%%%%%%%%%%%%%%%%%%%%%%%%%%%%%%%%%%%%%%%%%%%%%
\begin{abstract}

This paper presents a method to recover quadrotor \acp{UAV} from a throw, when no control parameters are known before the throw.
We leverage the availability of high-frequency rotor speed feedback available in racing drone hardware and software to find control effectiveness values and fit a motor model using recursive least squares (RLS) estimation.
Furthermore, we propose an excitation sequence that provides large actuation commands while guaranteeing to stay within gyroscope sensing limits. After 450ms of excitation, an \ac{INDI} attitude controller uses the 52 fitted parameters to arrest rotational motion and recover an upright attitude.
Finally, a \ac{NDI} position controller drives the craft to a position setpoint.
The proposed algorithm runs efficiently on microcontrollers found in common \ac{UAV} flight controllers, and was shown to recover an agile quadrotor every time in live experiments with as low as $3.5$m throw height, demonstrating robustness against initial rotations and noise.
We also demonstrate control of randomized quadrotors in simulated throws, where the parameter fitting \ac{RMS} error is typically within 10\% of the true value.

\end{abstract}

%%%%%%%%%%%%%%%%%%%%%%%%%%%%%%%%%%%%%%%%%%%%%%%%%%%%%%%%%%%%%%%%%%%%%%%%%%%%%%%%
\section{INTRODUCTION}

The number of different \acp{UAV} and their applications are rapidly increasing, and so are specific drone designs that fulfill various niches.
Traditionally, the development of control systems for drones requires system modeling and identification, control design and gain tuning, which takes valuable time and resources.
Yet, changes to the hardware configuration again require analysis, re-tuning and testing.
Ideally, the controller and its parameters would be identified automatically,
without the need for a simulation model or any manual analysis, design and testing.
Learning the control of a fixed wing aircraft model after a live drop from a balloon was demonstrated in \cite{heim_nasa_2018} with a global adaptive control approach.
To extend the current state of the art, the purpose of this study is to provide a method to control fixed-motor multirotor UAV using self-collected flight data, without prior knowledge of its flight characteristics. 

\begin{figure}[t]
    \centering
    \includegraphics[width=0.85\columnwidth]{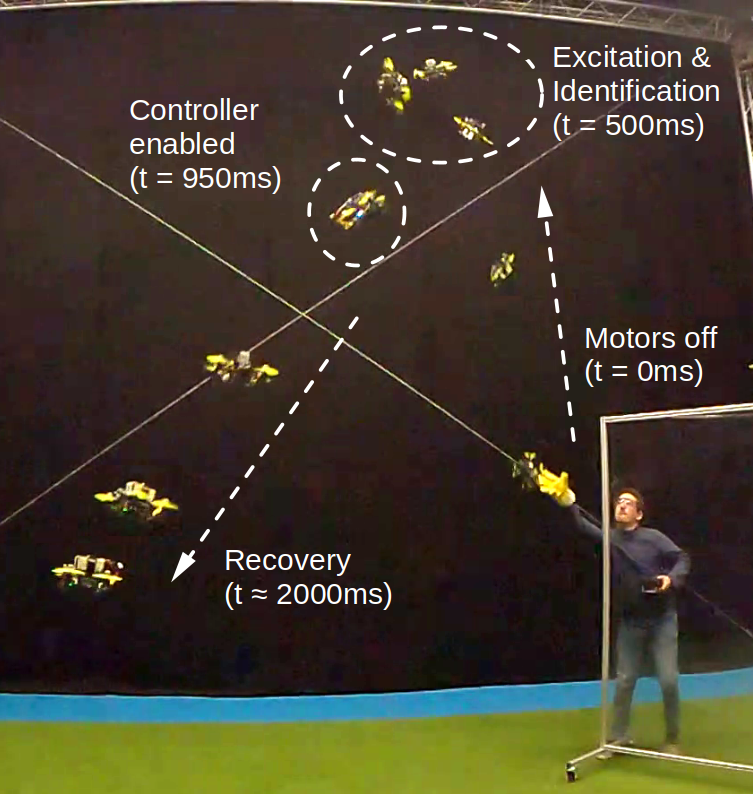}
    \caption{Composite image: The quadrotor learns to fly before it would hit the ground. Moving picture at \url{https://youtu.be/CLFPXcpzA14}}
    \label{fig:throwImage}
    \vspace*{-2mm}
\end{figure}

Reinforcement learning has been used to learn a quadrotor control policy from scratch \cite{eschmann_learning_2023}. This approach is applicable when a model of the quadrotor is already known, such that the learning can happen in simulation. The number of system interactions required to learn a satisfactory policy was $\approx 300,000$  ($\approx 1$h of simulated flight time). This is still far beyond what is safely and conveniently achievable without a simulation.

White-box adaptive multirotor controllers can be based on \ac{MRAC}, which have been shown to fly with an unknown swinging payload \cite{erasmus_robust_2020}. In this approach, the controller parameters are adapted directly to match a desired system response.
Given that quadrotors' performance limits may vary greatly, this reference model itself would likely also be adaptive.
Additionally, most \ac{MRAC} formulations assume knowledge of the signs in the control effectiveness. 

The problem therefor appears to lend itself better to controllers based on the physical properties and capabilities of the \ac{UAV}. Such controllers are also available and are often based on \ac{INDI}.
However, they are usually designed with the purpose to adapt to changes in the control effectiveness, after starting with a reasonable initial estimate \cite{smeur_adaptive_2016,li_nullspaceexcitationbased_2023}, or to recover from in-flight failures \cite{ke_uniform_2023}.

Insufficient excitation can have detrimental effect on estimation of the effectiveness for multirotors \cite{li_nullspaceexcitationbased_2023} and, since no prior knowledge is available for the problem at hand, a model-independent excitation method is required.
Furthermore, faster adaptation than achieved with \ac{LMS} in \cite{smeur_adaptive_2016} is needed, as well as to also identify an actuator model and not only effectiveness \cite{smeur_adaptive_2016,li_nullspaceexcitationbased_2023}.

In this work, we develop an algorithm that can learn to fly a multirotor directly from a single maneuver without prior platform knowledge, except the location and orientation of its \ac{IMU}, and that four upwards-pointing rotors are present.
We present (1) an attitude and position controller framework for multirotors, incorporating a novel linearization of the effects of rotor acceleration; (2) an excitation procedure that guarantees remaining within the gyroscope saturation limits; and (3) a \ac{RLS} estimator for the control effectiveness and motor model parameters.
It is demonstrated with parametric simulations and real-life throws, that the resulting algorithm reliably stabilizes and controls quadrotors after a throw of a few meters height, with unknown initial parameters (see figure \ref{fig:throwImage}). The algorithm computes entirely on onboard microprocessor hardware without simulation.

\section{METHODOLOGY}

\begin{figure*}[ht]
   \centering
   \includegraphics[width=\textwidth]{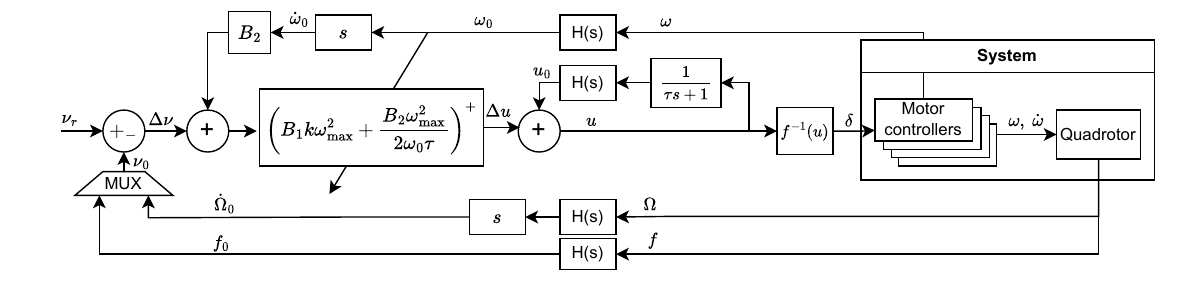}
   \caption{\ac{INDI} inner loop similar to \cite{smeur_adaptive_2016} with linearized motor acceleration dynamics}
   \label{fig:controller}
   \vspace*{-2mm}
\end{figure*}

This section describes control loops, parameter identification, and setup of the simulations and live experiments.

The quadrotor system is modeled as a control affine system perturbed by gravity, the thrust generated by the four rotors, and the reaction torque from actuating the rotors.

\subsection{Controller} \label{subsec:controller}

The flight controls used in the work can be broken down in 3 cascaded loops; (a) the ``inner loop'' controlling angular accelerations and specific forces, (b) the attitude controller, and (c) the position controller. They are described in detail in the following sections.

\paragraph{Inner Loop}

PID control is often used on the angular rate error $\Omega_{\text{ref}} - \Omega$. This generates torque commands which are then allocated to the motors in an open-loop fashion (``mixed'').
With \ac{NDI}, the relationship between motor states and angular acceleration may be modeled more accurately and is then inverted to obtain a relationship between so-called pseudo-controls $\nu$ and motor commands that linearises the dynamics with respect to the pseudo-controls. 
The authors of \cite{sieberling_robust_2010} show how \ac{INDI} can replace a part of the system model with online measurements.
This effectively closes the loop for angular acceleration and specific forces, and enables more responsive disturbance rejection and easier tuning, if the remaining model parameters (control effectiveness) are accurate. For an in-depth derivation, refer to \cite{sieberling_robust_2010,smeur_adaptive_2016}.

For flying vehicles, a common choice of pseudo-controls are the specific forces and angular accelerations  $\nu = \begin{pmatrix} f_x & f_y & f_z & \dot{p} & \dot{q} & \dot{r} \end{pmatrix}^T$, because a reference $\nu_r$ can be generated conveniently from attitude errors and body rotational velocity errors using linear error feedback.
For multirotors with rotor axes aligned with the local $z$-axis, 
$f_x=f_y=0$. To keep this work applicable for vehicles without this property or with different axes definitions however, we will only make this assumption when generating $v_r$.

An incremental model for $\Delta \nu$ has to capture the effects of the thrust increments $\Delta T\in \mathbb{R}^4$ produced by each of the motors, but also the rate of change of motor velocity $\dot\omega$, which exerts reaction torques on the vehicle.
This results in
\begin{align} \label{eq:INDI}
    \Delta \nu &= \nu - \nu_0 = B_1\ \Delta T + B_2\ \Delta \dot\omega\ ,
\end{align}
where $\nu_0$ is a measurement of $\nu$ from onboard sensors. $B_1$ and $B_2$ are called ``control effectiveness'' matrices and arise from the geometry and actuator parameters of the vehicle.

Neither $T$, $\omega$ nor $\dot\omega$ can be directly commanded with common \ac{ESC} hardware.
In the following, it is shown that the choice of thrust-normalized $u \triangleq T_s / T_{\text{max}}$ for the control variables enables a convenient linearized formulation suitable for control and identification. 
Only the term related to $\dot\omega$ is approximated, which only affects the yaw axis of multirotors.
Subscript $s$ stands for the steady-state values after cessation of actuator dynamics.

$u$ can be mapped to \ac{ESC} input $\delta \in [0, 1]$ by inverting a function $u=f(\delta)$, which in our work is assumed to be
\begin{align} \label{eq:ESCmodel}
    u = \left[\kappa \delta + (1-\kappa) \sqrt{\delta}\right]^2 \ ,
\end{align}
where $\kappa\geq 0$ is an unknown coefficient.
This matches motor test bench data and enables a parameter-linear model for identification, as shown in \ref{subsec:motormodel}.

Assume that the motor dynamics are described by a first-order lag, i.e. $\tau \dot \omega = \omega_s - \omega$ for some time constant $\tau > 0$ and steady-state motor speed $\omega_s$. The thrust produced by each rotor can be modeled with $T = k\omega^2$ \cite{smeur_adaptive_2016}, where $k$ is the propeller thrust constant.
Combining these two models yields
\begin{align}
    \tau\dot\omega + \omega &= \omega_s = \sqrt{\frac{T_s}{k}} = \sqrt{\frac{T + \Delta T_s}{k}} = \sqrt{\frac{T}{k}}\sqrt{1+\frac{\Delta T_s}{T}}\label{eq:motor_model}\ .
\end{align}

For small steady-state thrust increments $\Delta T_s$, the square root can be eliminated using the binomial approximation. Additionally, substituting $\omega = \sqrt{T/k}$ and $\Delta u = \Delta T_s / T_\text{max} = \Delta T_s / k\omega_\text{max}^2$ gives
\begin{align}
    \tau\dot\omega + \omega &\approx \sqrt{\frac{T}{k}}\left(1+\frac{\Delta T_s}{2T}\right) = \omega\left(1+\frac{\Delta T_s}{2T}\right) \\
    \tau\dot\omega &\approx \frac{\omega}{2T}\Delta T_s = \frac{1}{2k\omega} \Delta T_s = \frac{\omega_\text{max}^2}{2\omega} \Delta u\ .
\end{align}

With these approximations, we can write a version of (\ref{eq:INDI}) that is linear in the non-dimensional thrust increment $\Delta u$
\begin{align}
    \nu - \nu_0 &= B_1\Delta T_s + B_2 \left( \dot\omega - \dot\omega_0\right) \\
    \nu - \nu_0 + B_2 \dot\omega_0 &\approx \left( B_1 k\omega_\text{max}^2 + \frac{B_2 \omega_\text{max}^2}{2\omega_0\tau} \right) \Delta u \label{eq:INDIcontrol}\ .
\end{align}
Given a reference $\nu = \nu_r$, this can be solved using e.g. the pseudo-inverse or with active-set methods, if bounds on $u$ must be satisfied \cite{blaha_survey_2023}.
If $k$ is known, then $u_0$ may be calculated from a measurement of $\omega$. Otherwise, emulating the actuator dynamics $U_0 = \frac{U}{\tau s + 1}$ has shown success, even though this lag actually acts on the rotational speed and not the thrust.

The subscript $0$ need not imply the values of the previous time-step. To combat the effects of noise, amplified by the differentiation of $\omega$ and $\Omega$, we can replace each $\dot\Omega_0$, $u_0$, and $\nu_0$ with low-pass filtered versions.
A suitable choice for $H(s)$ is a second-order Butterworth \cite{bacon_reconfigurable_2001}, which has a steeper cutoff compared to a first-order filter, at the expense of larger phase-shift. However, the larger phase delay does not affect the system dynamics since all values are filtered with the same filter, preserving the validity of (\ref{eq:INDI}). The filtering however does decrease disturbance rejection performance \cite{smeur_adaptive_2016}.
A cutoff of $f_c = 15$Hz yielded good performance in our experiments, but lower values should be used if structural vibrations are a concern, or if smoother motor commands are desired.

Figure \ref{fig:controller} shows a diagram of the inner loop.

\paragraph{Attitude control}

An attitude setpoint given by unit quaternion $q_r$ may be tracked by calculating an error quaternion in a body-fixed frame $q_e^B = q^{-1} q_r$ where $q$ is the current attitude.
Reference angular body velocities can then be generated with 
\begin{align}\label{eq:attCtl}
    \Omega_r = 2\arccos{\left((q_e^B)_w\right)} \left( \hat n \circ A\right)\ ,
\end{align}
where $\hat n$ is the normalized vector part of $q_e^B$, subscript $w$ denotes the scalar part of $q_e^B$, $A$ is a vector of gains, and $\circ$ denotes element-wise multiplication.
The last 3 elements of $\nu_r$ in (\ref{eq:INDIcontrol}) are set to $D\circ(\Omega_r - \Omega)$, where $\Omega$ is the measured angular body velocity and $D$ another vector of 3 gains.

\paragraph{Position control}

Attitude and thrust are related to inertial acceleration through:
\begin{align} \label{eq:posNDI}
    a^I = R^I_B f^B + G^I\ ,
\end{align}
in the absence of aerodynamic effects (including wind). Here, $f^B = \begin{pmatrix} 0&0&f_z^B \end{pmatrix}^T$ are the specific forces generated by the motors in a body-fixed frame, $G^I$ is the gravity vector, and $R^I_B$ the rotation from the body-fixed to the inertial frame.
PID control is used to generate inertial acceleration references $a_r^I$ from position and velocity estimates.

If a reference yaw-angle $\Psi$ is prescribed, then (\ref{eq:posNDI}) can be solved explicitly for $R^{I,*}_B$ and $f_z^{B,*}$, yielding an \ac{NDI} law. $R^{I,*}$ is the setpoint for the attitude controller, and $\nu_{r,1:3} = \begin{pmatrix}0&0&f_z^{B,*}\end{pmatrix}^T$.

\paragraph{Gain tuning}

With perfect modeling of $B_1 k \omega_\text{max}^2$ and $B_2$, and in the absence of outside disturbances, it is shown in \cite{smeur_adaptive_2016} that the entire inner loop collapses to a first-order transfer function coincident with the actuator model.
The inner loop angular rate gains $D$ may then be tuned by pole-placement based on a known actuator time constant $\tau$ and a prescribed damping ratio $\zeta_D$.
By approximating the resulting 2nd order system by a 1st order linear lag $D / (s + D)$, the pole-placement technique can be successively applied to yield gains for attitude, velocity and position. We choose $\zeta_D = 0.8$, $\zeta_A = 0.7$, $\zeta_V = 0.7$, $\zeta_P = 0.9$.

\subsection{Model Identification} \label{subsec:identification}

To summarize, the controller depends on the system parameters $B_1 k$, $B_2$, $\omega_\text{max}^2$, $\tau$ and the mappings $u = f(\delta)$.
This section shows a method to identify them from online flight data using recursive least squares estimation (RLS).

\paragraph{RLS}

For parameter-linear models $Y = X^T \Theta$, with scalar $Y$, regressor column vector $X$, and parameter column vector $\Theta$, a common recursive estimator is \ac{RLS} \cite{madisetti_digital_2009}: 
\begin{align} % added indices k
    e_k &\gets Y_k - X_k^T\Theta_{k-1}\\
    K_k &\gets P_{k-1}X_k(\lambda + X_k^T P_{k-1} X_k)^{-1} \label{eq:RLSgain}\\
    \Theta_k &\gets \Theta_{k-1} + K_k e_k\\
    P_k &\gets \lambda^{-1}(P_{k-1} - K_k X_k^T P_{k-1})\ ,\label{eq:RLScov}
\end{align}
where $P_0 = I\cdot 10^2$, $\Theta_0 = 0$.

Since our \ac{RLS} filters described in the subsequent sections have to compute on single precision floating-point hardware, we avoid the accumulation of numerical errors by pre-scaling the regressors and outputs to be in the order of unity.
The gain $K_k$ is decreased, if the subtraction in (\ref{eq:RLScov}) would otherwise accrue errors. 

Finally, choosing $\lambda = \exp{(P_s/0.2)}$, where $P_s$ is the sampling period, ensures that the error observations $e_k$ decay by $63$\% after $0.2$ seconds, which gave satisfactory performance in the short identification time.
To avoid divergence of $P_k$, its entries are limited to $10^{10}$, but future research should replace this by an adaptive forgetting procedure instead.

\paragraph{Motor model} \label{subsec:motormodel}

Typical \acp{ESC} drive the rotors to $\omega_\text{idle} > 0$, even when $\delta=0$.
While the thrust produced at that speed has been neglected in (\ref{eq:ESCmodel}), the unknown offset needs to be included when working with rotor speed data.
With $\omega = \sqrt{T/k}$, $u \triangleq T_s/T_\text{max}$: 
\begin{align}
    \omega_s = \omega_\text{max} \sqrt{u} + \omega_\text{idle} = \omega_\text{max} \left(\kappa \delta + (1-\kappa) \sqrt{\delta} \right) + \omega_\text{idle}\ .
\end{align}

$\omega_\text{max}$ depends on battery state of charge and aerodynamic flow conditions, but this is neglected as $\omega_\text{max}$ will be estimated online.

To turn this into a parameter-linear formulation, the parameters $a,b>0$ are introduced and defined by the transformations $\omega_\text{max} = a + b$ and $\kappa = \frac{a}{a+b}$. Combined with the first-order motor dynamics model $\tau \dot \omega = \omega_s - \omega$ used above, this yields
\begin{align}
    \omega &= a \delta + b \sqrt{\delta} + \omega_\text{idle} - \tau\dot\omega \label{eq:motor_model_ident}.
\end{align}

\paragraph{Effectiveness $B_1$ and $B_2$}

To identify $B_1$ and $B_2$, we again turn to (\ref{eq:INDI}) and notice that the only non-measurable quantity is $\Delta T$.
If we write the propeller thrust model $T=k\omega^2$ in linearized differential form $\Delta T = 2k\cdot \omega\ \Delta\omega$, we obtain 
\begin{align} \label{eq:INDIident}
    \nu - \nu_0 = \Delta\begin{pmatrix}f_x \ f_y \ f_z\ \dot p\ \dot q \ \dot r \end{pmatrix}^T = B_1 k \left( 2 \omega\ \Delta\omega\right) + B_2 \Delta \dot\omega\ ,
\end{align}
where only $B_1 k$ and $B_2$ are unknown and all other quantities can be trivially derived from measurements of specific force $f$, body rotation rate $\Omega$ and motor rotation rate $\omega$. Modern flight control hardware and software is available that provides even the motor rotation rate $\omega$ and $\dot\omega$ at sufficiently high rates.
Again, we can low-pass filter each quantity, in this case chosen as a $20$Hz second-order Butterworth low-pass.

The first and last 3 rows in (\ref{eq:INDIident}) each form a linear estimation problem in 8 parameters. However, since there is no conceivable way in which $\Delta\dot\omega$ gives rise to forces, its term is neglected for the first 3 rows, such that they only require 4 parameters.
The problem appears to require $6$ separate RLS filters, but since the first and last 3 rows share the same regressors, the expensive operations (\ref{eq:RLSgain}) and (\ref{eq:RLScov}) only have to be computed once for each group of filters.

\paragraph{Excitation}

For the parameters above to converge using RLS, sufficient excitation must be present. 
A unique challenge with multirotors is that even a short burst ($<100$ms) of extreme motor input can induce rotations that exceed the $\pm 2000^\circ$/s measurement range of most MEMS gyroscopes.
This has to be avoided, as attitude estimation would be lost making recovery impossible. On the other extreme, too conservative excitation may be drowned in measurement noise.
Our approach aims at inducing large accelerations on the vehicle without saturating the gyroscope, by providing an adaptive sequence of excitations to the motors. Feedback control is enabled only after excitation (and identification) is complete.

\begin{figure}[ht]
   \centering
   \includegraphics[width=\columnwidth]{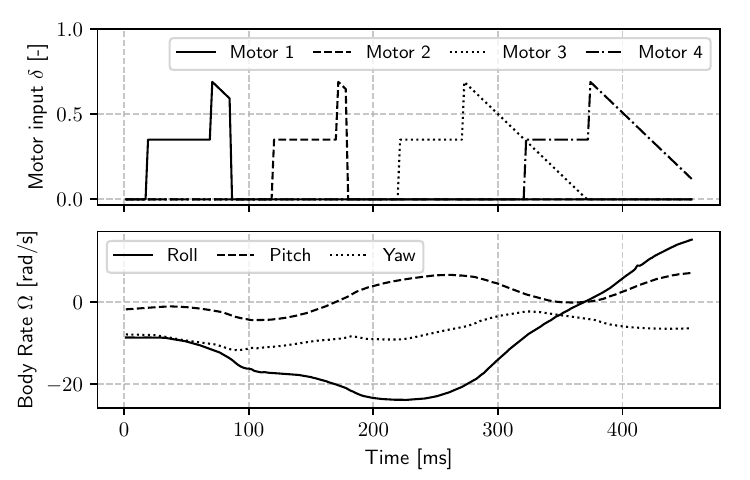}
   \caption{Excitation after detection of throwing. In this trial, the commands for motors 1 and 2 are truncated to avoid sensor saturation in the roll axis.}
   \label{fig:excitation}
   \vspace*{-2mm}
\end{figure}

Each motor is successively excited with two step inputs followed by a decreasing ramp. This ensures that many different levels of $\delta$, and the largest possible $\dot\omega$ are provided to the system.
On entry to the first step of each motor, the gyroscope reading $\Omega_e$ and margin $\Delta \Omega$ to the sensor limits is recorded. The excitation is aborted if a gyroscope reading exceeds $\Omega_e + \Delta \Omega / n_l$ where $n_l$ is the number of motors left to excite after the current motor. This is illustrated in figure \ref{fig:excitation}.
It could be argued that this is too conservative, since if a quadrotor is able to hover, then 2 of its motors must provide a rotation that de-saturates the sensor, so $0.5n_l$ in the denominator should be sufficient.
However, if we assume that rotor acceleration and deceleration have the same dynamics, then in the worst case, the thrust from a rotor imparts as much angular impulse on spin-up as it does while spinning down after the excitation is aborted.
This effectively halves the permissible margin $\Delta \Omega$.

It seems generally accepted that orthogonal sine-inputs are more sample efficient inputs for the identification of MIMO systems \cite{morelli_practical_2021}.
Furthermore, it is common to mix the control signal with the generated signal \cite{heim_nasa_2018}. However, without prior knowledge of the vehicles capabilities, it seems challenging to guarantee the safety of these inputs while ensuring sufficient excitation.

\begin{figure*}[t]
   \centering
   \vspace*{3mm}
   \includegraphics[width=\textwidth, trim={0cm, 0.2cm, 0cm, 0.7cm}, clip]{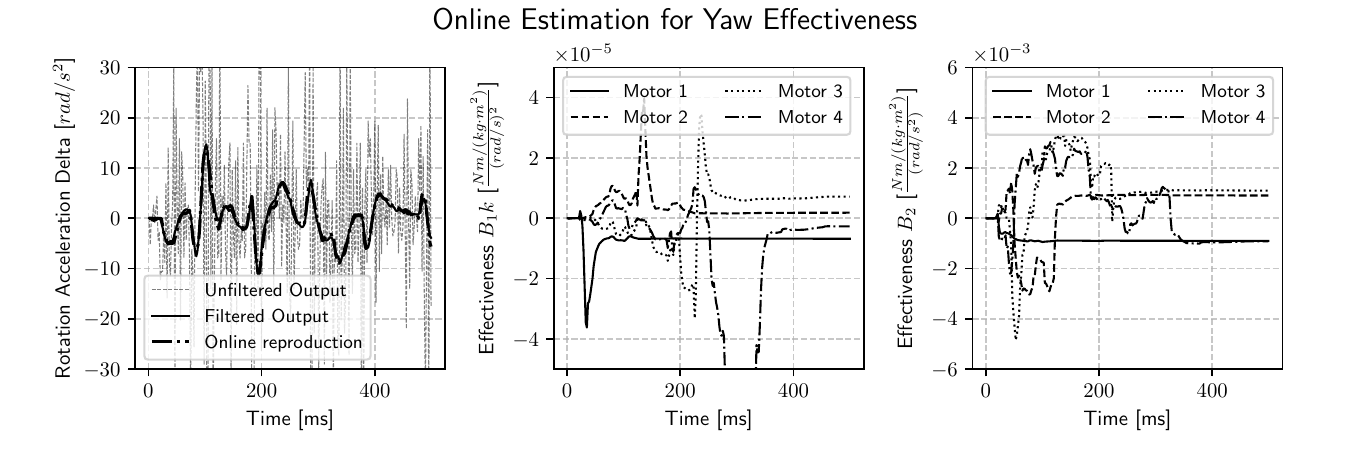}
   \caption{Time evolution of the control effectiveness around the yaw axis. }
   \label{fig:fx_estimation_r}
   \vspace*{-3mm}
\end{figure*}

\begin{figure}[t]
   \centering
   \includegraphics[width=\columnwidth]{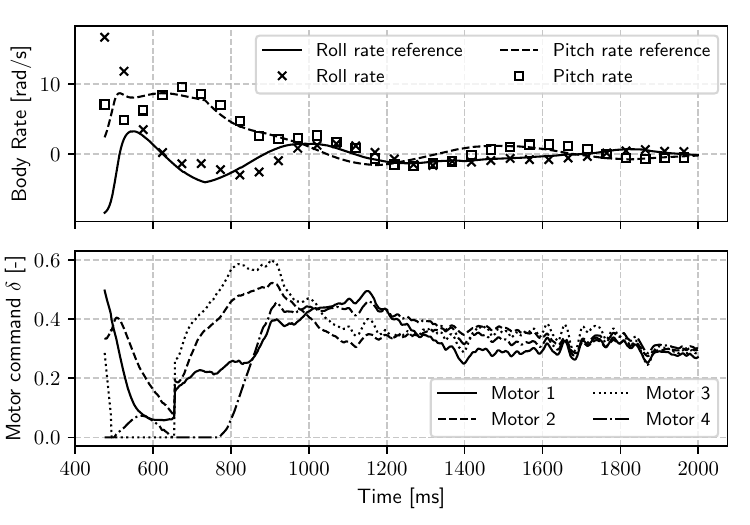}
   \caption{Angular rate tracking and motor commands during recovery}
   \label{fig:recovery}
   \vspace*{-2mm}
\end{figure}

\subsection{Experimental Setup} \label{subsec:experimentsetup}

A 3-inch-propeller quadrotor UAV, sporting an STM32H743 microcontroller and TDK InvenSense ICM-42688-P IMU was used for experimental verification.
The identification and control algorithm above was implemented onboard at 2kHz sample rate in the open-source software package Betaflight\footnote{\url{https://betaflight.com/}}. Its focus on performance and computational efficiency, and especially the availability of motor rotation rate feedback at equally high sample rate, make it ideal to research low-level control implementations.
Our adaptations were so significant that a new fork INDIflight\footnote{\url{https://github.com/tudelft/indiflight}} was created.

The UAV was instrumented with reflective markers to enable optical position tracking. The positioning data was sent to the UAV, but only used for position control; it does not influence identification.
Aside from the positioning triangulation and transmitting that data to the drone, all calculations are performed onboard.
The internal attitude estimation remains the Mahony filter provided by Betaflight. Heading, velocity and position use simple constant gain observers to fuse inertial measurements with optical position and velocity data. %If less accurate or reliable positioning systems like GNSS are to be used for future outdoor experiments, these estimators would likely have to be replaced by an Extended Kalman Filter.

\paragraph{Conditions}

The nominal \ac{INDI} controller parameters were calculated with bench tests of propeller and motors, and measurements of geometry and mass properties. 
For controlled experiments, this controller took the quadrotor from ground level onto a $3.5$-$4$m high parabolic trajectory and induced a pseudo-random rotation between 0 and 10rad/s.
Using the known geometry of the quadrotor, and after performing bench tests with a controller-motor-propeller combination, the parameters were calculated.
The engines are then spooled down for 250ms, and the excitation is triggered. After its completion 450ms later, the fitting is done and switchover to the identified parameter-set occurs.

%When manually throwing, the release is detected using the accelerometer, and a longer delay time of 500ms is used, for safety of the experimenter.

\subsection{Simulation Setup} \label{subsec:simulationsetup}

$1000$ asymmetrical multirotor crafts were generated and simulated in randomized throwing conditions of about $4$ height. For each rotor of each craft, we varied the following parameters:
motor assignment;
distance from CoG $[0.05,\ 0.15]m$;
top-view angle of the CoG-to-rotor line from a nominal square quadrotor $[-30^\circ,\ +30^\circ]$;
non-linearity $\kappa \in [0.25,\ 0.75]$;
time constant $\tau \in [15,\ 35]ms$;
drag-moment coefficient $[0.01,\ 0.02]$. If the parameters resulted in a craft that is physically unable to hover, it was regenerated.

The simulation is linked against the same INDIflight software as used in the experiments, and can be reproduced using Docker\footnote{\href{https://github.com/tudelft/indiflightSupport/tree/iros_imav_2024}{github.com/tudelft/indiflightSupport/tree/iros\_imav\_2024}}.
All data files are also available\footnote{\href{https://doi.org/10.4121/0530be90-cc6c-4029-9774-670657882906}{10.4121/0530be90-cc6c-4029-9774-670657882906}}.

%The open-source robot simulator Gazebo is used to obtain parameterizable quadrotor simulations. The motor locations, time constant, moment constant and input non-linearity $\kappa$ were drawn from uniform distributions, while ensuring that the resulting vehicle is still physically capable of hover.
%
%A plugin based on the Open Software Robotics Foundation's \texttt{vehicle\_gateway} project\footnote{\url{github.com/osrf/vehicle_gateway}} was then used to run 16 parameter-randomized simulations with the flight controller in the loop.
%The initial conditions were a parabolic trajectory of approximately $4$m height. $450$ms after the launch, the excitation sequence is triggered, and after that the described controller drives the vehicle to a point $1.5$m above the origin.

\section{RESULTS}

\begin{table*}[t!]
\begin{center}
\vspace{3mm}
\caption{Mean and standard deviation for $B_1$, $B_2$ and the motor model during 20 launches with different initial conditions.} \label{tab:parametersExperiment}
\begin{tabular}{|l||rrr|rrr|rrr|rrr|}
\hline
& \multicolumn{3}{c|}{Motor 1} & \multicolumn{3}{c|}{Motor 2} & \multicolumn{3}{c|}{Motor 3} & \multicolumn{3}{c|}{Motor 4}\\
 & Reference & Mean & Std & Reference & Mean & Std & Reference & Mean & Std & Reference & Mean & Std \\
\hline
\hline
$B_{1,x}\cdot k \cdot 10^6$ & 0.000 & 0.056 & 0.066 & 0.000 & -0.101 & 0.034 & 0.000 & 0.068 & 0.064 & 0.000 & -0.082 & 0.046 \\
$B_{1,y}\cdot k \cdot 10^6$ & 0.000 & -0.089 & 0.046 & 0.000 & 0.012 & 0.039 & 0.000 & 0.007 & 0.049 & 0.000 & -0.019 & 0.042 \\
$B_{1,z}\cdot k \cdot 10^6$ & -0.621 & -0.658 & 0.060 & -0.621 & -0.592 & 0.030 & -0.621 & -0.792 & 0.069 & -0.621 & -0.684 & 0.043 \\
$B_{1,p}\cdot k \cdot 10^6$ & -23.440 & -29.551 & 3.194 & -23.440 & -30.661 & 2.360 & 23.440 & 29.904 & 2.952 & 23.440 & 27.441 & 2.287 \\
$B_{1,q}\cdot k \cdot 10^6$ & -15.718 & -18.992 & 2.040 & 15.718 & 21.283 & 1.498 & -15.718 & -14.013 & 1.136 & 15.718 & 18.492 & 1.277 \\
$B_{1,r}\cdot k \cdot 10^6$ & -2.989 & -6.587 & 0.673 & 2.989 & 2.319 & 0.480 & 2.989 & 6.669 & 0.792 & -2.989 & -2.158 & 0.303 \\
\hline
$B_{2,p}\cdot 10^3$ & 0.000 & -0.002 & 0.079 & -0.000 & 0.185 & 0.038 & -0.000 & 0.168 & 0.079 & 0.000 & -0.040 & 0.074 \\
$B_{2,q}\cdot 10^3$ & 0.000 & 0.066 & 0.049 & -0.000 & -0.030 & 0.057 & -0.000 & 0.070 & 0.125 & 0.000 & 0.057 & 0.084 \\
$B_{2,r}\cdot 10^3$ & -1.011 & -0.915 & 0.021 & 1.011 & 0.945 & 0.035 & 1.011 & 1.061 & 0.052 & -1.011 & -0.961 & 0.054 \\
\hline
$\omega_{\text{max}}$ & 4113 & 5253 & 248.2 & 4113 & 5278 & 199.7 & 4113 & 4812 & 58.3 & 4113 & 4919 & 81.6 \\
$\kappa$ & 0.460 & 1.118 & 0.088 & 0.460 & 1.184 & 0.079 & 0.460 & 0.994 & 0.008 & 0.460 & 1.000 & 0.009 \\
$\omega_{\text{idle}}$ & 450 & 516.7 & 7.306 & 450 & 512.7 & 8.100 & 450 & 505.8 & 7.194 & 450 & 503.7 & 7.271 \\
$\tau$ [ms] & 20.00 & 25.91 & 0.598 & 20.00 & 26.21 & 0.793 & 20.00 & 26.64 & 0.425 & 20.00 & 26.61 & 0.294 \\
\hline
\end{tabular}
\end{center}
\vspace*{-2mm}
\end{table*}

\begin{table}[ht]
\caption{Parameter error \ac{RMS} over the $1000$ simulated runs.} \label{tab:parametersSim}
\begin{tabular}{|l||r|rrrr|}
\hline
 & Ground truths & \multicolumn{4}{c|}{\ac{RMS} error}\\
 & mean-absolute &  Motor 1 &  2 &  3 &  4 \\
\hline
\hline
$B_{1,x}\cdot k\cdot 10^6$ & 0.000 & 0.037 & 0.036 & 0.032 & 0.037 \\
$B_{1,y}\cdot k\cdot 10^6$ & 0.000 & 0.038 & 0.034 & 0.032 & 0.036 \\
$B_{1,z}\cdot k\cdot 10^6$ & 0.463 & 0.040 & 0.036 & 0.033 & 0.039 \\
$B_{1,p}\cdot k\cdot 10^6$ & 18.301 & 1.147 & 1.515 & 1.597 & 1.379 \\
$B_{1,q}\cdot k\cdot 10^6$ & 17.538 & 1.107 & 1.518 & 1.535 & 1.424 \\
$B_{1,r}\cdot k\cdot 10^6$ & 4.173 & 0.472 & 0.436 & 0.395 & 0.395 \\
\hline
$B_{2,p}\cdot 10^3$ & 0.000 & 0.055 & 0.068 & 0.085 & 0.114 \\
$B_{2,q}\cdot 10^3$ & 0.000 & 0.055 & 0.064 & 0.089 & 0.116 \\
$B_{2,r}\cdot 10^3$ & 0.667 & 0.030 & 0.032 & 0.039 & 0.052 \\
\hline
$\omega_{\text{max}}$ & 4900 & 390.7 & 275.7 & 148.7 & 47.46 \\
$\kappa$ & 0.499 & 0.119 & 0.105 & 0.114 & 0.101 \\
$\omega_{\text{idle}}$ & 0.000 & 2.824 & 5.568 & 11.28 & 4.660 \\
$\tau$ [ms] & 25.04 & 0.627 & 0.689 & 0.768 & 1.080 \\
\hline
\end{tabular}
\vspace*{-2mm}
\end{table}

\subsection{Experiments}

A total of 57 controlled launches were performed, and during all of them the parameters were identified well enough to stabilize the UAVs attitude. Due to an oversight, suitable log files were only obtained in the last 20 runs.

Figure \ref{fig:fx_estimation_r} shows the time evolution of the 8 parameters in the model for $\Delta \dot r$ during an experiment run (the 4 parameters in the center plot relate to each motor velocity and 4 parameters in the right plot relate to motor acceleration). The low-pass filtering is effective in reducing noise in the measured targets shown in the left plot.
From the center and right plot it can be seen that for all motors but motor 1 the parameters initially do not converge. This is due to the presence of only idle rotation speed before the step input. However, once excitation is provided, the parameters immediately converge, and reproduction of the rotation acceleration measurement is unaffected by the initial uncertainty.
Nonetheless, control based on these parameters would be impossible.

Despite initial rate errors in excess of $20$rad/s, the inner loop tracks the rate reference after less than $200$ms, with relatively smooth motor commands, as seen in figure \ref{fig:recovery}. Slight oscillations are present between $1200$ and $2000$ms, which may indicate that the choices of $\zeta$ have been too low.

Finally, table \ref{tab:parametersExperiment} shows the means and standard deviations of the fitted parameters over the last 20 runs. They are compared to the calculated ``reference'' value from bench tests, geometry and mass properties measurements. Note, that these calculations also have their own uncertainty and should not be seen as a ground truth.
In particular, it is reasonable that motors 1 and 3 (the rear motors) have higher yaw effectiveness $B_{1,r}$ and higher thrust effectiveness $B_{1,z}$ than motors 2 and 4, because the structure that the motors are mounted on is much wider in the front of our quadrotor.
This can reduce the downwards flow velocity and swirl.

Any effects from non-zero inflow are neither captured on the static test bench, nor in the simulation. It is therefore possible, that the differences in the motor model parameters in the experiments are due to these effects occurring in-flight.
Consequently, the motor model (\ref{eq:motor_model_ident}) and likely the thrust model $T = k\omega^2$ do not hold globally.

\subsection{Simulations}

All $1000$ vehicles were successfully stabilized despite errors in the parameter estimates.
Table \ref{tab:parametersSim} shows that the root-mean-square (RMS) errors between the true and estimated parameters over the runs are typically below $10$\%.
The motor numbers $1$ to $4$ correspond to the order in which they have been excited.
The ``nominal'' column gives context to interpret the RMS values, and is the mean of the absolute parameter value over all the true generated crafts.

%%%%%%%%%%%%% Conclusions and backmatter %%%%%%%%%%%%%%%%%%%%%%%%%

\section{CONCLUSIONS and FUTURE WORK}

The simulation and real-world experiments show that the classical recursive parameter estimation method described, combined with a short and simple excitation sequence lasting only $450$ms, is enough to recover agile quadrotor UAVs when thrown to heights as low as $3.5$m.
It identifies all $52$ required control parameters of quadrotor UAV and tunes $4$ gains well enough for recovery and position control. In terms of sample efficiency, this beats current reinforcement learning methods by orders of magnitude and does not require simulations.
In terms of computational efficiency, the proposed scheme runs at a rate of 2kHz on consumer flight control hardware.

Future research may aim at eliminating dependencies, such as the high frequency motor velocity feedback, and the knowledge of IMU location and orientation.
Furthermore, physical experiments with other multirotor designs would validate the generalizability of the proposed method, and expansions to other types of vehicles should be investigated.
Different excitation sequences may increase sample efficiency, and keeping the adaptation active during recovery and subsequent flight may reduce time for recovery, increase controller performance and enable adaptation to different flight conditions.


%%%%%%%%%%%%%%%%%%%%%%%%%%%%%%%%%%%%%%%%%%%%%%%%%%%%%%%%%%%%%%%%%%%%%%%%%%%%%%%%

%%%%%%%%%%%%%%%%%%%%%%%%%%%%%%%%%%%%%%%%%%%%%%%%%%%%%%%%%%%%%%%%%%%%%%%%%%%%%%%%

%%%%%%%%%%%%%%%%%%%%%%%%%%%%%%%%%%%%%%%%%%%%%%%%%%%%%%%%%%%%%%%%%%%%%%%%%%%%%%%%

%\section*{APPENDIX}

\section*{ACKNOWLEDGMENT}

We would like to thank Coen C. de Visser for his helpful suggestions and proofreading.

%%%%%%%%%%%%%%%%%%%%%%%%%%%%%%%%%%%%%%%%%%%%%%%%%%%%%%%%%%%%%%%%%%%%%%%%%%%%%%%%
% References

%\bibliography{IEEEtranBST/IEEEabrv,IROSLearn2Fly}   % uncomment to use bibtex

                         % or input the bibtex-generated bbl

\end{document}